\documentclass{article}

     \usepackage[final]{neurips_2019}

\usepackage[utf8]{inputenc} 
\usepackage[T1]{fontenc}    
\usepackage{hyperref}       
\usepackage{url}            
\usepackage{booktabs}       
\usepackage{amsfonts}       
\usepackage{nicefrac}       
\usepackage{microtype}      
\usepackage{xcolor}
\usepackage{amsthm}
\usepackage{amsmath}
\usepackage{enumitem}

\usepackage{hyperref}
\usepackage{url}
\usepackage{commath}
\usepackage{nicefrac}
\usepackage{graphicx}
\usepackage{caption}
\usepackage{subcaption}
\usepackage{makecell}

\usepackage{amssymb}
\usepackage{amsmath}
\usepackage{amsthm}
\usepackage{amsfonts}
\usepackage[super]{nth}
\usepackage{bbm}

\usepackage{commath}
\usepackage{amssymb}

\usepackage{amsfonts}
\usepackage{graphicx,wrapfig,lipsum}

\usepackage{xparse}

\usepackage[utf8]{inputenc} 
\usepackage[T1]{fontenc}    
\usepackage{booktabs}       
\usepackage{nicefrac}       
\usepackage{microtype} 

\title{Non-Gaussianity of Stochastic Gradient Noise}

\author{%
  Abhishek Panigrahi\\
  Microsoft Research, Bangalore, India \\
  \texttt{t-abpani@microsoft.com} \\
  \And
  Raghav Somani \\
  University of Washington, Seattle, WA, USA \\
  \texttt{raghavs@cs.washington.edu} \\
  \AND
  Navin Goyal \\
  Microsoft Research, Bangalore, India \\
  \texttt{navingo@microsoft.com} \\
  \And
  Praneeth Netrapalli \\
  Microsoft Research, Bangalore, India \\
  \texttt{praneeth@microsoft.com} \\
}

\DeclareMathOperator*{\argmin}{arg\,min}

\begin{document}

\maketitle

\begin{abstract}
  What enables Stochastic Gradient Descent (SGD) to achieve better generalization than Gradient Descent (GD) in Neural Network training? This question has attracted much attention. In this paper, we study the distribution of the Stochastic Gradient Noise (SGN) vectors during the training. We observe that for batch sizes $256$ and above, the distribution is best described as Gaussian at-least in the early phases of training. This holds across data-sets, architectures, and other choices.
\end{abstract}

\theoremstyle{plain}
\newtheorem{theorem}{Theorem}[section]
\newtheorem{corollary}{Corollary}[theorem]
\newtheorem{lemma}[theorem]{Lemma}
\newtheorem{proposition}[theorem]{Proposition}

\theoremstyle{definition}
\newtheorem{definition}[theorem]{Definition}
\newtheorem{remark}[theorem]{Remark}

\newcommand{\round}[1]{\left( #1 \right)}
\newcommand{\curly}[1]{\left\lbrace #1 \right\rbrace}
\newcommand{\squarebrack}[1]{\left\lbrack #1 \right\rbrack}

\newcommand{\sumi}[2]{\sum\limits_{i=#1}^{#2}}
\newcommand{\sumj}[2]{\sum\limits_{j=#1}^{#2}}
\newcommand{\sumk}[2]{\sum\limits_{k=#1}^{#2}}
\newcommand{\sump}[2]{\sum\limits_{p=#1}^{#2}}
\newcommand{\suml}[2]{\sum\limits_{l=#1}^{#2}}
\newcommand{\sumn}[2]{\sum\limits_{n=#1}^{#2}}
\newcommand{\summ}[2]{\sum\limits_{m=#1}^{#2}}
\newcommand{\sumt}[2]{\sum\limits_{t=#1}^{#2}}

\newcommand{\Sum}{\sum_{i = 1}^{n}}
\newcommand{\Sumi}[1]{\sum\limits_{i = 1}^{#1}}
\newcommand{\Sumt}[1]{\sum\limits_{t = 1}^{#1}}


\newcommand{\esqnorm}[1]{\left\lVert#1\right\rVert_2^2}
\newcommand{\enorm}[1]{\left\lVert#1\right\rVert_2}
\newcommand{\infnorm}[1]{\left\lVert#1\right\rVert_\infty}
\newcommand{\opnorm}[1]{\left\lVert#1\right\rVert_\text{op}}
\newcommand{\normF}[1]{\left\lVert#1\right\rVert_{\text{F}}}
\newcommand{\inner}[1]{\left\langle#1\right\rangle}
\newcommand{\ceil}[1]{\left\lceil#1\right\rceil}
\newcommand{\floor}[1]{\left\lfloor#1\right\rfloor}

\newcommand{\zero}{\mathbf{0}}
\newcommand{\one}{\mathbf{1}}

\newcommand{\avec}{\mathbf{a}}
\newcommand{\bvec}{\mathbf{b}}
\newcommand{\cvec}{\mathbf{c}}
\newcommand{\dvec}{\mathbf{d}}
\newcommand{\e}{\mathbf{e}}
\newcommand{\f}{\mathbf{f}}
\newcommand{\g}{\mathbf{g}}
\newcommand{\h}{\mathbf{h}}
\newcommand{\ivec}{\mathbf{i}}
\newcommand{\jvec}{\mathbf{j}}
\newcommand{\kvec}{\mathbf{k}}
\newcommand{\lvec}{\mathbf{l}}
\newcommand{\m}{\mathbf{m}}
\newcommand{\n}{\mathbf{n}}
\newcommand{\ovec}{\mathbf{o}}
\newcommand{\p}{\mathbf{p}}
\newcommand{\q}{\mathbf{q}}
\newcommand{\rvec}{\mathbf{r}}
\newcommand{\s}{\mathbf{s}}
\newcommand{\tvec}{\mathbf{t}}
\newcommand{\uvec}{\mathbf{u}}
\newcommand{\vvec}{\mathbf{v}}
\newcommand{\w}{\mathbf{w}}
\newcommand{\x}{\mathbf{x}}
\newcommand{\y}{\mathbf{y}}
\newcommand{\z}{\mathbf{z}}

\newcommand{\A}{\mathbf{A}}
\newcommand{\B}{\mathbf{B}}
\newcommand{\C}{\mathbf{C}}
\newcommand{\D}{\mathbf{D}}
\newcommand{\Emat}{\mathbf{E}}
\newcommand{\F}{\mathbf{F}}
\newcommand{\G}{\mathbf{G}}
\newcommand{\Hmat}{\mathbf{H}}
\newcommand{\I}{\mathbf{I}}
\newcommand{\J}{\mathbf{J}}
\newcommand{\K}{\mathbf{K}}
\newcommand{\Lmat}{\mathbf{L}}
\newcommand{\M}{\mathbf{M}}
\newcommand{\N}{\mathbf{N}}
\newcommand{\Omat}{\mathbf{O}}
\newcommand{\Pmat}{\mathbf{P}}
\newcommand{\Q}{\mathbf{Q}}
\newcommand{\Rmat}{\mathbf{R}}
\newcommand{\Smat}{\mathbf{S}}
\newcommand{\T}{\mathbf{T}}
\newcommand{\U}{\mathbf{U}}
\newcommand{\V}{\mathbf{V}}
\newcommand{\W}{\mathbf{W}}
\newcommand{\X}{\mathbf{X}}
\newcommand{\Y}{\mathbf{Y}}
\newcommand{\Z}{\mathbf{Z}}

\newcommand{\SIGMA}{\mathbf{\Sigma}}
\newcommand{\LAMBDA}{\mathbf{\Lambda}}

\newcommand{\Acal}{\mathcal{A}}
\newcommand{\Bcal}{\mathcal{B}}
\newcommand{\Ccal}{\mathcal{C}}
\newcommand{\Dcal}{\mathcal{D}}
\newcommand{\Ecal}{\mathcal{E}}
\newcommand{\Fcal}{\mathcal{F}}
\newcommand{\Gcal}{\mathcal{G}}
\newcommand{\Hcal}{\mathcal{H}}
\newcommand{\Ical}{\mathcal{I}}
\newcommand{\Jcal}{\mathcal{J}}
\newcommand{\Kcal}{\mathcal{K}}
\newcommand{\Lcal}{\mathcal{L}}
\newcommand{\Mcal}{\mathcal{M}}
\newcommand{\Ncal}{\mathcal{N}}
\newcommand{\Ocal}{\mathcal{O}}
\newcommand{\Pcal}{\mathcal{P}}
\newcommand{\Qcal}{\mathcal{Q}}
\newcommand{\Rcal}{\mathcal{R}}
\newcommand{\Scal}{\mathcal{S}}
\newcommand{\Tcal}{\mathcal{T}}
\newcommand{\Ucal}{\mathcal{U}}
\newcommand{\Vcal}{\mathcal{V}}
\newcommand{\Wcal}{\mathcal{W}}
\newcommand{\Xcal}{\mathcal{X}}
\newcommand{\Ycal}{\mathcal{Y}}
\newcommand{\Zcal}{\mathcal{Z}}

\newcommand{\alphavec}{\boldsymbol{\alpha}}
\newcommand{\betavec}{\boldsymbol{\beta}}
\newcommand{\gammavec}{\boldsymbol{\gamma}}
\newcommand{\deltavec}{\boldsymbol{\delta}}
\newcommand{\epsvec}{\boldsymbol{\epsilon}}
\newcommand{\etavec}{\boldsymbol{\eta}}
\newcommand{\nuvec}{\boldsymbol{\nu}}
\newcommand{\tauvec}{\boldsymbol{\tau}}
\newcommand{\rhovec}{\boldsymbol{\rho}}
\newcommand{\lmbda}{\boldsymbol{\lambda}}
\newcommand{\muvec}{\boldsymbol{\mu}}
\newcommand{\thetavec}{\boldsymbol{\theta}}

\newcommand{\BigO}[1]{\mathcal{O}\round{#1}}
\newcommand{\BigOmega}[1]{\Omega\round{#1}}

\newcommand{\R}{\mathbb{R}}
\newcommand{\Rd}[1]{\mathbb{R}^{#1}}
\newcommand{\Natural}{\mathbb{N}}
\newcommand{\Integer}{\mathbb{Z}}
\newcommand{\Rational}{\mathbb{Q}}

\newcommand{\E}[1]{\mathbb{E}\squarebrack{#1}}
\newcommand{\Exp}[2]{\mathbb{E}_{#1}\squarebrack{#2}}
\newcommand{\Prob}[1]{P\curly{#1}}

\newcommand{\inv}[1]{\frac{1}{#1}}
\newcommand{\indicator}[2]{\mathbbm{1}_{#1}\round{#2}}
\newcommand{\Tr}[1]{\text{Tr}\squarebrack{#1}}

\newcommand{\BOX}[1]{\fbox{\parbox{\linewidth}{\centering#1}}}
\newcommand{\textequal}[1]{\stackrel{#1}{=}}
\newcommand{\textleq}[1]{\stackrel{#1}{\leq}}
\newcommand{\textgeq}[1]{\stackrel{#1}{\geq}}
\newcommand{\defeq}{\vcentcolon=}

\newcommand{\dd}[2]{\frac{d #1}{d #2}}
\newcommand{\ddn}[3]{\frac{d^{#1} #2}{d #3^{#1}}}
\newcommand{\dodo}[2]{\frac{\partial #1}{\partial #2}}

\section{Introduction}
Stochastic Gradient Descent (SGD) algorithm~\cite{robbins1951stochastic} and its variants are workhorses of modern deep learning, e.g. \cite{Bottou_Nocedal, goodfellow2016deep}. Not only does SGD allow efficient training of Neural Networks (a non-convex optimization problem) achieving small training loss, it also achieves good generalization, often better compared to Gradient Descent (GD). While striking progress has been made in theoretical understanding of highly over-parameterized networks in recent years, for realistic settings we lack satisfactory understanding. In particular, what property of the distribution of the small-batch gradients arising in SGD is responsible for the efficacy of SGD?

Considerable effort has been expended on understanding SGD for Neural Networks. In particular, there have been attempts to relate SGD to a discretization of a continuous time process via Stochastic Differential Equations (SDEs), e.g., \cite{mandt2016variational,raginsky2017non,zhang2017hitting,hu2017diffusion,jastrzkebski2017three,zhu2018anisotropic}. The Stochastic Gradient (SG) can be written as Gradient + Stochastic Gradient Noise (SGN).  \cite{mandt2016variational,hu2017diffusion,jastrzkebski2017three,zhu2018anisotropic} assume that SGN is approximately Gaussian when the batch size $b$ is large enough, assuming the Central Limit Theorem (CLT) conditions, which we discuss in the next section.  \cite{raginsky2017non,zhang2017hitting} add explicit gaussian noise to stochastic gradients in each iteration. One can analyze these algorithms by approximating them with the continuous time process in the form of Langevin diffusion, and one can show that the Markov process exhibited by the continuous time iterates is ergodic with its unique invariant measure whose log-density is proportional to the negative of the objective function~\cite{roberts2002langevin}. Therefore, SGD can be seen as a first-order Euler-Maruyama discretization of the Langevin dynamics~\cite{jastrzkebski2017three,li2017stochastic}.

Since the gradient vectors are very high-dimensional, the number of samples required in the application of the CLT, so that a multivariate Gaussian distribution provides a sufficiently accurate approximation, may be very high, raising doubts about the applicability of the CLT for SGN computed using \textit{small} batch size.
For SGD training of neural networks, the assumption of SGD noise being Gaussian-like was contested recently in \cite{tail-index}. They argue that the SGN instead follows a stable distribution with infinite variance obtained by the application of the Generalized Central Limit Theorem (GCLT). They provided experimental evidence for their claim. Furthermore, they provided arguments based on the properties of the associated continuous time process which can then be viewed as an SDE with the stochastic term being a \textit{Levy motion}.

\section{Preliminaries}\label{sec:prelims}
The problem of training a Neural Network is essentially an unconstrained optimization problem that usually constitutes a finite sum non-convex objective function. Function $f: \mathbb{R}^p \times \mathbb{R}^{d+1}  \to \mathbb{R}$ denotes the composition of the Neural Network and the loss function. On 
learnable parameters $\w\in\R^p$, and input $(\x, y) \in \mathbb{R}^d \times \mathbb{R}$, function value $f(\w; (\x, y))$ measures the accuracy of the underlying Neural Network on $(\x, y)$. For a set of $n$ training data samples $\left\{\x_i, y_i\right\}_{i=1}^{n} \in \left(\mathbb{R}^{d} \times \mathbb{R}\right)^{n}$, define $f\round{\w; B} := \inv{\abs{B}}\sum_{i\in B}f\left(\w; \left\{\x_i, y_i\right\} \right)$, and $\nabla_\w f\round{\w; B} := \inv{\abs{B}}\sum_{i\in B} \nabla_w f(\w; \{\x_i, y_i\} )$. Then the optimization problem is given by
\begin{equation}
\w^{*} \in \argmin_{\w\in\R^p} f\left(\w; \squarebrack{n} \right).
\end{equation}
SGD update rule is given by 
$\w^{(t+1)} \gets \w^{(t)} - \eta \nabla_{\w} f(\w^{(t)}, B_t),$
where $\w^{(t)}$ denotes the iterate at time $t\geq 0$, $\eta>0$ denotes the learning rate, and $B_t\subseteq{\squarebrack{n}}$ denotes a batch of training examples of size $b \le n$ picked at time $t$. A random SGN vector at time $t$ is then given by $f(\w^{(t)}; B_t) - f(\w^{(t)}; [n])$.

\paragraph{Gaussian random variables.}
Perhaps not as well-known as the scalar CLT, there is a CLT for vector-valued random variables showing that the appropriately scaled sums of i.i.d. random variables with finite covariance matrix tends to multivariate Gaussian in distribution.
A useful fact about multivariate Gaussians is that if for a random vector $\mathbf{X}$, all its one-dimensional marginals are univariate Gaussians then
$\mathbf{X}$ must be multivariate Gaussian (see, e.g., Exercise 3.3.4 in \cite{vershynin_book}).

\paragraph{Stable distribution.} For a real-valued random variable $X$, let $X_1$ and $X_2$ denote its i.i.d. copies. $X$ is said to be \textit{stable} if for any $a, b > 0$, random variable $a X_1 + b X_2$ has the same distribution as $cX+ d$ for some constants $c > 0$ and $d$. Stable distributions are a four parameter family $S(\alpha, \beta, \gamma, \delta)$ of univariate probability distributions. There is no closed form formula known for these distributions in general, instead they are defined using their characteristic function. We do not include that definition here since we will not need it. Instead, we will just note the properties of stable distributions that are relevant for us. See~\cite{nolan} for more information. The special case of $S(\alpha, 0, 1, 0)$ is represented as $\mathcal{S}\alpha\mathcal{S}$ distribution, and can be considered as the symmetric, centered and normalized case of stable distribution.
$\alpha \in (0, 2]$ is the stability parameter; we omit the discussion of other parameters. This family includes the standard Gaussian distribution ($S(2, 0, 1, 0)$), and standard Cauchy distribution ($S(1, 0, 1, 0)$).
Interest in stable distributions comes from the {\bf Generalized Central Limit Theorem} (see~\cite{nolan}) which informally states that the distribution of appropriately scaled sums of i.i.d. random variables with infinite variance tends to a stable distribution. This was the basis of the suggestion in~\cite{tail-index} that SGN vectors follow a stable distribution with infinite variance. More specifically,~\cite{tail-index} posited that the coordinates of SGN vectors are i.i.d. stable random variables. In parallel with multivariate Gaussians, there is a notion of multivariate stable distributions (see~\cite{nolan}) which satisfies the property that one-dimensional marginals are univariate stable. A more general hypothesis for SGN vectors could be that they are multivariate stable.

\section{Method, Experiments and Observations}
Gaussianity testing is a well-studied problem in statistics; we use tests due to~\cite{Shapiro} and~\cite{Anderson}. These were found to be the top two tests in a comparison study~\cite{Razali}. Both of the tests assume the null hypothesis that the given set of scalar examples come from a Gaussian distribution.
A small $p$-value implies that the null hypothesis is likely to be false. We follow the following steps to perform statistical tests on stochastic gradient noises (SGN).

\begin{enumerate}[leftmargin=*]
	\item  We train a model using minibatch SGD and consider SGN vectors for $1000$ independent minibatches, which are of the same size as the training batch-size and randomly sampled with replacement, at multiples of $100$ iterations.
	\item We project each SGN vector along $1000$ random unit vectors, and perform Gaussianity tests on the projections along each direction. We plot the average confidence value across the dimensions, as given by Shapiro--Wilk test~\cite{Shapiro}, and  the fraction of directions accepted by the Anderson Wilk test~\cite{Anderson} and compare with the case when actual Gaussian data is fed to the tests.
\end{enumerate}

\subsection{Experimental Setup}
We conduct experiments on different models, namely a $3$-layer fully connected network, Alexnet, Resnet18 and VGG16 model, with and without batch-normalization (BN), and different data-sets, namely CIFAR10 and MNIST~\cite{CIFAR,MNIST,Krizhevsky,He,Simonyan}. All the layers were initialized with Xavier initialization~\cite{glorot2010understanding}. The models were trained with constant learning rate mini-batch SGD, where mini-batch-size was varied from $32$, $256$ and $4096$, on cross entropy loss. The learning rates were varied in $10^{\curly{-1,-2,-3}}$.

\subsection{Observations and Conclusion}

\begin{figure} \label{fig:Resnet18-256}
	\centering
	\begin{subfigure}{.30\textwidth}
		\centering
		\includegraphics[width=\linewidth]{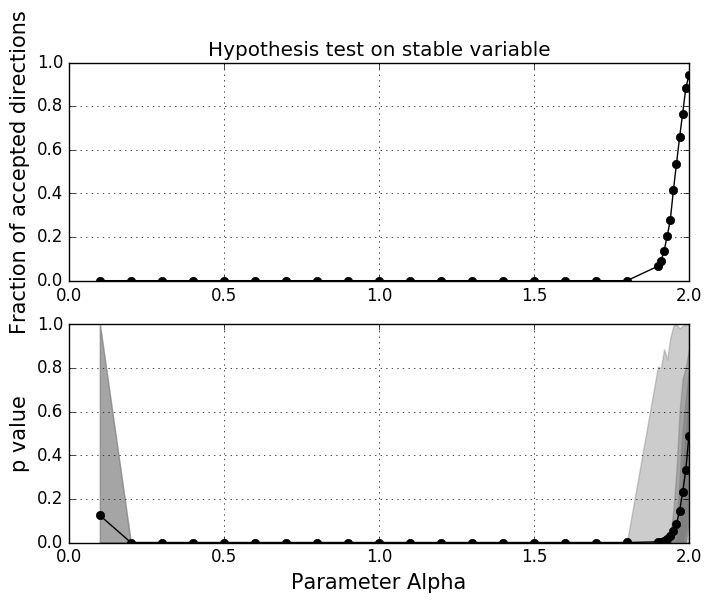}
		\caption{}
		\label{fig:Levy_test}
	\end{subfigure}
	\begin{subfigure}{.37\textwidth}
		\centering
		\includegraphics[width=\linewidth]{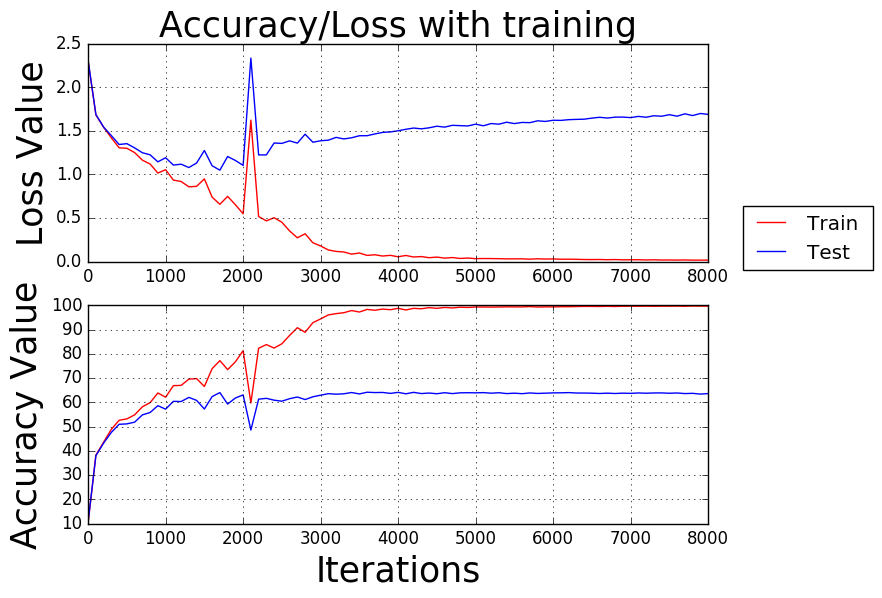}
		\caption{}
		\label{fig:Resnet_withBN_lr1e-2_256bt_lossacc}
	\end{subfigure}
	\begin{subfigure}{.30\textwidth}
		\centering
		\includegraphics[width=\linewidth]{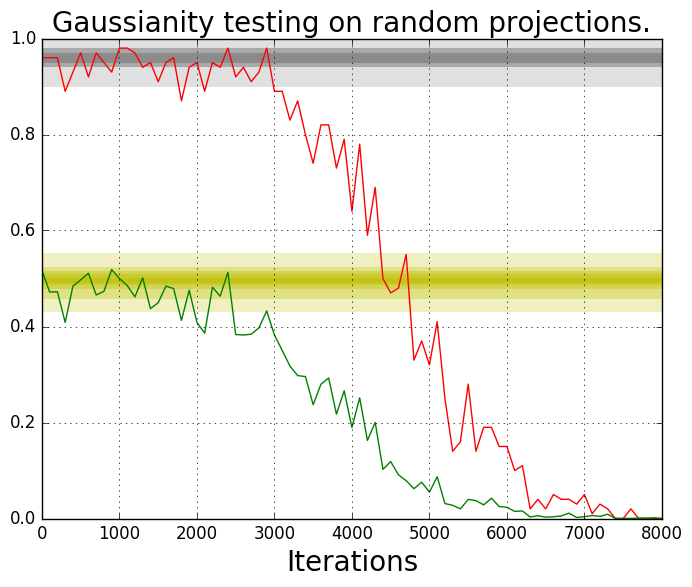}
		\caption{}
		\label{fig:Resnet_woBN_lr1e-2_256bt_projtest}
	\end{subfigure}
	
	\begin{flushright}
		\begin{subfigure}{.60\textwidth}
			\includegraphics[width=\linewidth]{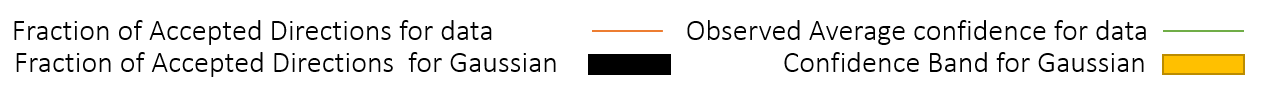}
		\end{subfigure}
	\end{flushright}
	\caption{(a) Gaussianity tests on $\mathcal{S}\alpha\mathcal{S}$ variables; (b) and (c) training Resnet18 with $256$ batch-sized SGD at l.r. $10^{-2}$, (b) Accuracy \& loss curves, and (c) Gaussianity tests on projection.} 
\end{figure}

We first perform sanity checks to see how well the hypothesis tests do on generated $\mathcal{S}\alpha\mathcal{S}$ samples. It can be seen in~\autoref{fig:Levy_test} that for $0.2\lesssim \alpha \lesssim 1.8$ the tests very confidently and correctly declare that the distribution is not Gaussian. As $\alpha \to 2$, we expect the distribution to become closer to Gaussian, and for very small $\alpha$ the tails become very heavy. Thus the behavior of the statistical tests on $\mathcal{S}\alpha\mathcal{S}$ distribution is as expected.

\begin{figure}
	\centering
	\begin{subfigure}{.22\textwidth}
		\centering
		\includegraphics[width=\linewidth]{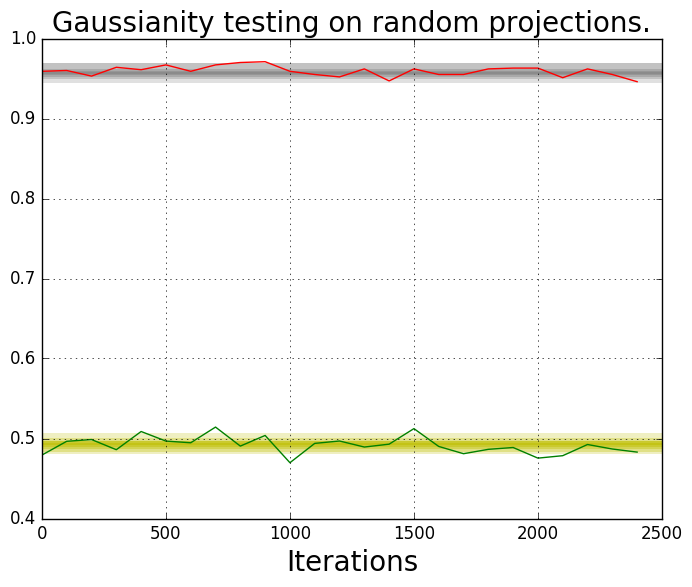}
		\caption{}
		\label{fig:Resnet_withBN_lr1e-2_4096bt_projtest}
	\end{subfigure}
	\begin{subfigure}{.27\textwidth}
		\centering
		\includegraphics[width=\linewidth]{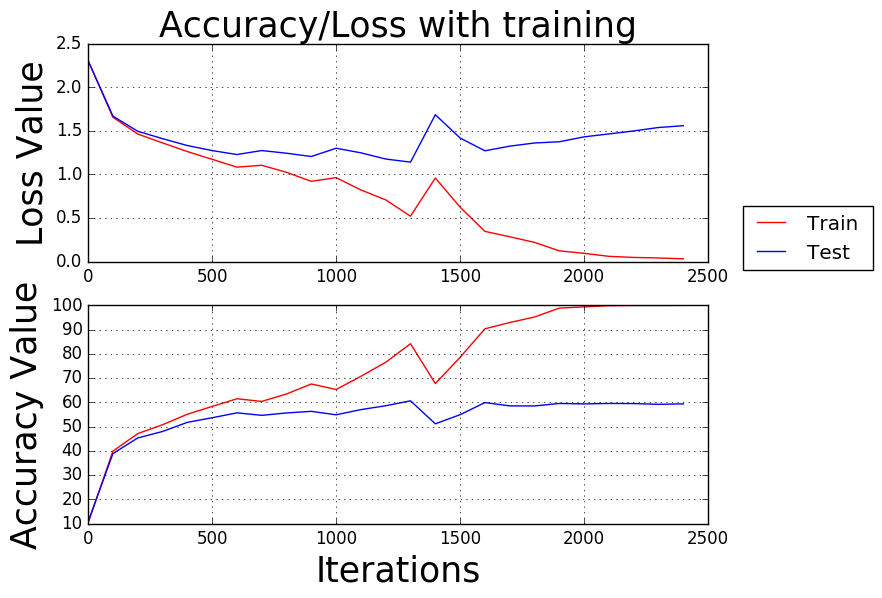}
		\caption{}
		\label{fig:Resnet_withBN_lr1e-2_4096bt_lossacc}
	\end{subfigure}
	\begin{subfigure}{.22\textwidth}
		\centering
		\includegraphics[width=\linewidth]{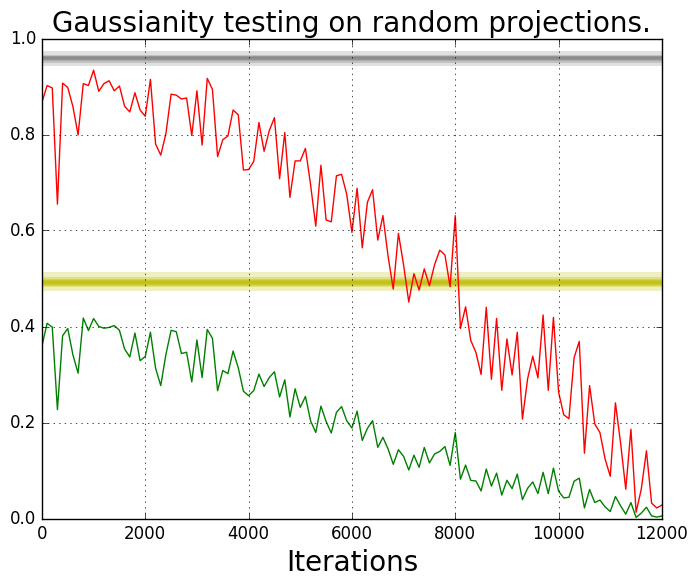}
		\caption{}
		\label{fig:Resnet_wthBN_lr1e-2_32bt_projtest}
	\end{subfigure}
	\begin{subfigure}{.27\textwidth}
		\centering
		\includegraphics[width=\linewidth]{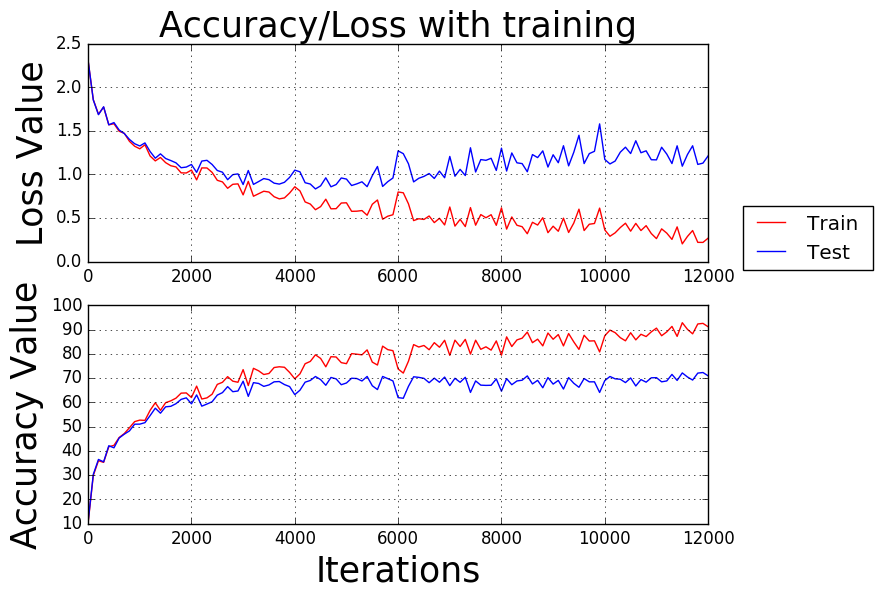}
		\caption{}
		\label{fig:Resnet_withBN_lr1e-2_32bt_lossacc}
	\end{subfigure}
	\caption{(a) and (b) training Resnet18 with $4096$ batch-sized SGD at l.r. $10^{-2}$; (a) Gaussianity tests on projections (b) Accuracy \& loss behavior with training; (c) and (d) training Resnet18 with $32$ batch-sized SGD at l.r. $10^{-2}$; (c) Gaussianity tests on projections (d) Accuracy \& loss curves. For (a) and (c), legend is the same as the legend for Fig.~\ref{fig:Resnet_woBN_lr1e-2_256bt_projtest}.}
\end{figure}

We see that SGN looks Gaussian to statistical tests in Fig~\ref{fig:Resnet_withBN_lr1e-2_4096bt_projtest} for Resnet18 throughout the training for batch-size $4096$. In contrast, in Fig.~\ref{fig:Resnet_wthBN_lr1e-2_32bt_projtest} the behavior is not Gaussian anywhere for batch-size $32$. For intermediate batch-sizes such as $256$, behavior was Gaussian at the beginning of the training, and later becomes non-Gaussian. We observe this behavior consistently across different architectures, data-sets, and larger batch-sizes. This is in contrast with the results in~\cite{tail-index}, which suggests stable distribution right from the start for batch-size $500$. When the behavior is Gaussian in $1000$ random directions, by the characterization of multivariate Gaussian mentioned above, this suggests---though does not definitively prove---that SGN vectors are multivariate Gaussians early on in the training. For batch-size $32$ is it possible that the SGN distribution here is stable? This is unlikely because by GCLT we would expect this behavior at higher batch-sizes, but we observe Gaussian behavior.

We briefly explain possible reasons why our results do not agree with those in~\cite{tail-index}. They use the estimator of \cite{Mohammadi} to estimate $\alpha$. The application of this estimator in \cite{tail-index} suffers from 2 errors: (1) In~\cite{Mohammadi}, the estimator is shown to give a good estimate of $\alpha$ under the assumption that the random variable is indeed $\alpha$-stable; it makes no claims about the estimate when this assumption is not satisfied. \cite{tail-index} seems to tacitly assume stability. (2) The estimator assumes that the coordinates of SGN vector are i.i.d. This assumption is invalid in the typical over-parameterized setting.

\newpage
\bibliographystyle{plain}
\bibliography{references}
\newpage
\appendix
\section{Appendix}
Here we present the various plots for the remaining data-sets and models.

\begin{figure}[!h]
	\centering
	\begin{subfigure}{.32\textwidth}
		\centering
		\includegraphics[width=\linewidth]{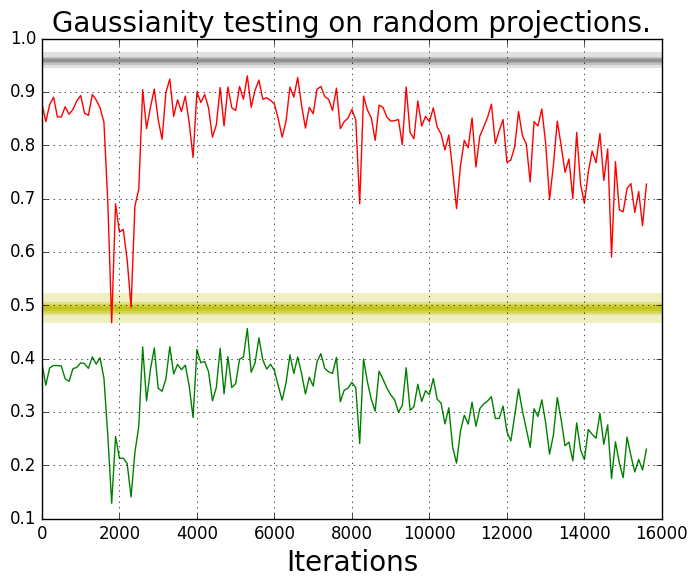}
		\caption{}
	\end{subfigure}
	\begin{subfigure}{.32\textwidth}
		\centering
		\includegraphics[width=\linewidth]{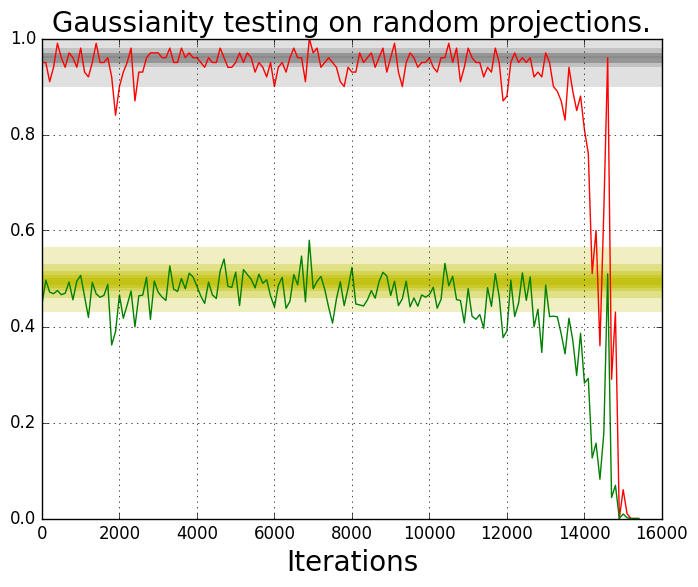}
		\caption{}
	\end{subfigure}%
	\begin{subfigure}{.32\textwidth}
		\centering
		\includegraphics[width=\linewidth]{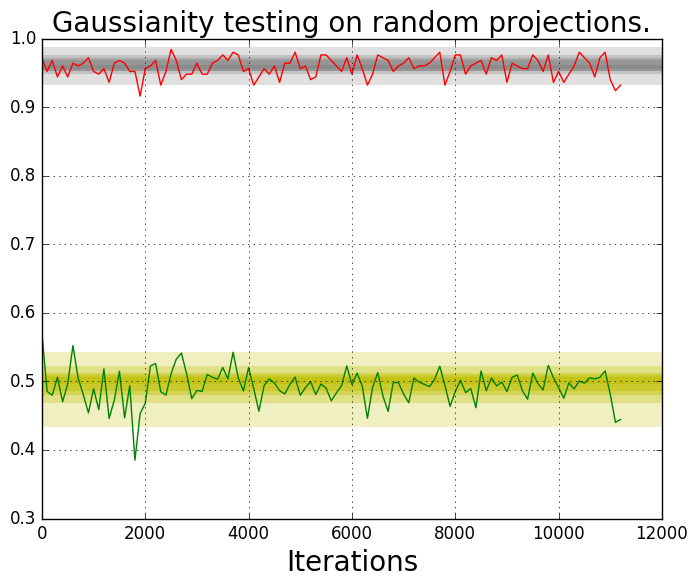}
		\caption{}
	\end{subfigure}

	\begin{flushright}
		\begin{subfigure}{.60\textwidth}
			\includegraphics[width=\linewidth]{Legend_images.png}
		\end{subfigure}
	\end{flushright}
	\caption{Gaussianity Test Experiments on VGG16, without BN and trained on CIFAR10 at l.r. $10^{-2}$, (a) Mini batch-size is $32$ (b) Mini batch-size is $256$ (c) Mini batch-size is $4096$. } 
	
\end{figure}

\begin{figure}[!h]
	\centering
	\begin{subfigure}{.32\textwidth}
		\centering
		\includegraphics[width=\linewidth]{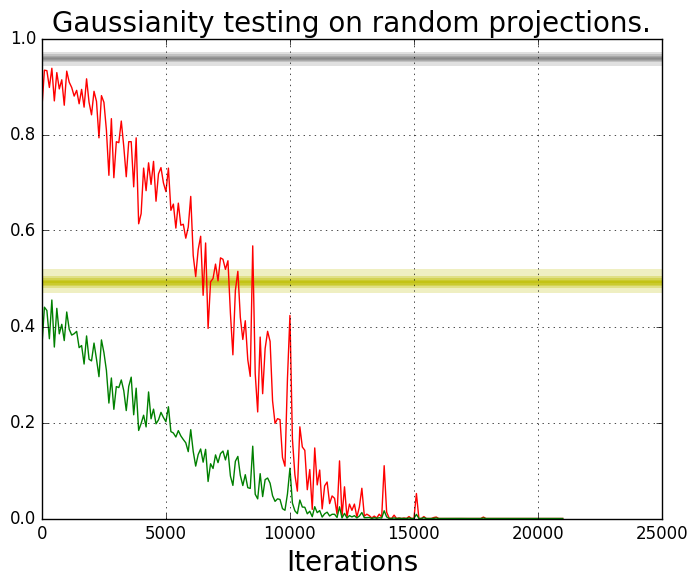}
		\caption{}
	\end{subfigure}
	\begin{subfigure}{.32\textwidth}
		\centering
		\includegraphics[width=\linewidth]{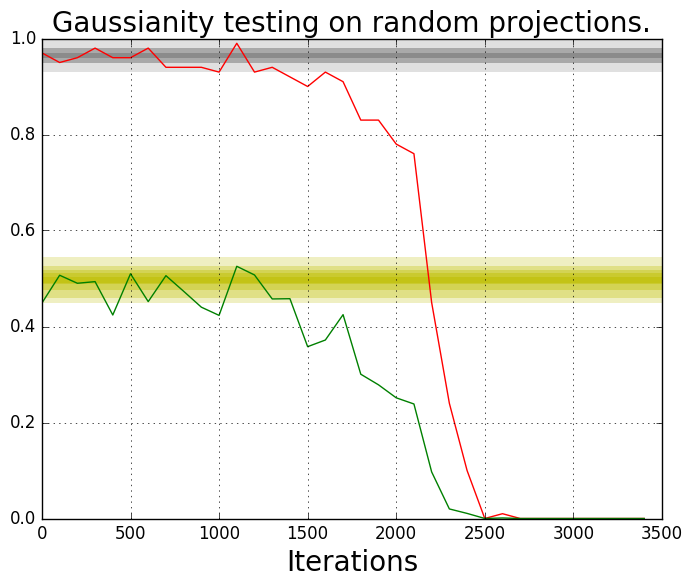}
		\caption{}
	\end{subfigure}%
	\begin{subfigure}{.32\textwidth}
		\centering
		\includegraphics[width=\linewidth]{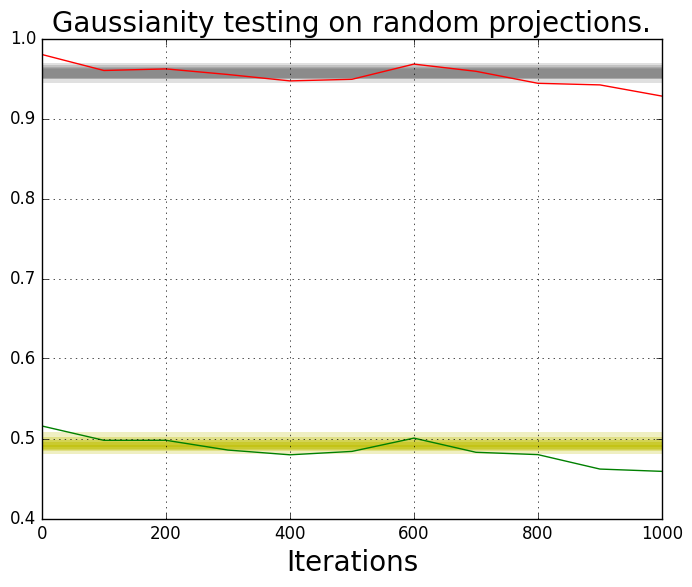}
		\caption{}
	\end{subfigure}

	\begin{flushright}
		\begin{subfigure}{.60\textwidth}
			\includegraphics[width=\linewidth]{Legend_images.png}
		\end{subfigure}
	\end{flushright}
	\caption{Gaussianity Test Experiments on VGG16, with BN and trained on CIFAR10 at l.r. $10^{-2}$, (a) Mini batch-size is $32$ (b) Mini batch-size is $256$ (c) Mini batch-size is $4096$.} 
	
\end{figure}

\begin{figure}
	\centering
	\begin{subfigure}{.32\textwidth}
		\centering
		\includegraphics[width=\linewidth]{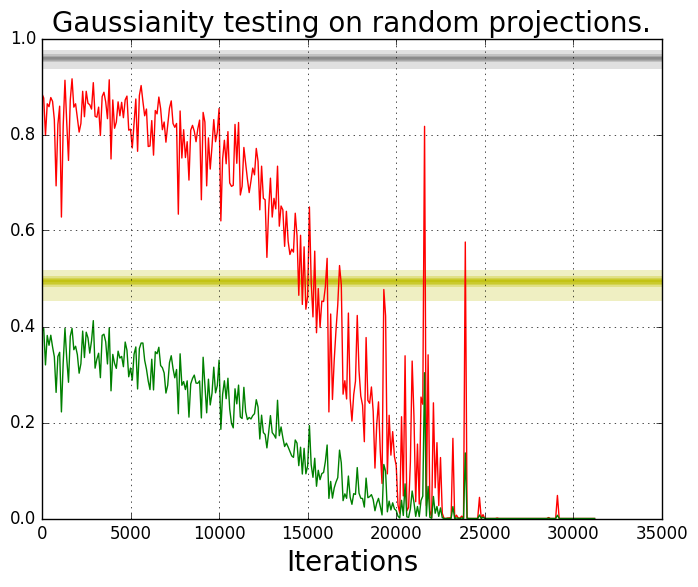}
		\caption{}
	\end{subfigure}
	\begin{subfigure}{.32\textwidth}
		\centering
		\includegraphics[width=\linewidth]{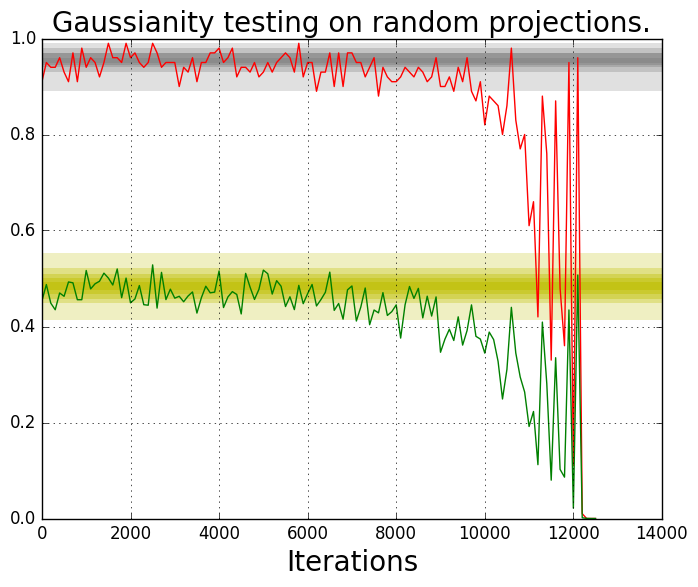}
		\caption{}
	\end{subfigure}%
	\begin{subfigure}{.32\textwidth}
		\centering
		\includegraphics[width=\linewidth]{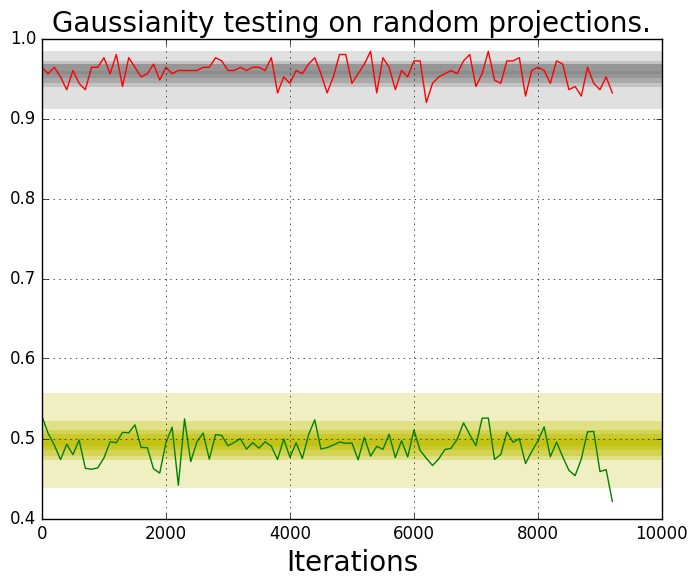}
		\caption{}
	\end{subfigure}

	\begin{flushright}
		\begin{subfigure}{.60\textwidth}
			\includegraphics[width=\linewidth]{Legend_images.png}
		\end{subfigure}
	\end{flushright}
	\caption{Gaussianity Test Experiments on Resnet18, without BN and trained on CIFAR10 at l.r. $10^{-2}$, (a) Mini batch-size is $32$ (b) Mini batch-size is $256$ (c) Mini batch-size is $4096$.} 
	
\end{figure}
\begin{figure}
	\centering
	\begin{subfigure}{.32\textwidth}
		\centering
		\includegraphics[width=\linewidth]{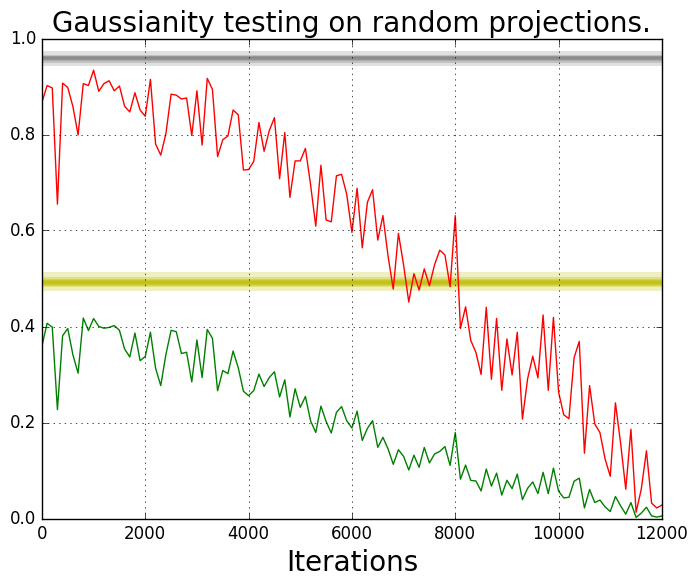}
		\caption{}
	\end{subfigure}
	\begin{subfigure}{.32\textwidth}
		\centering
		\includegraphics[width=\linewidth]{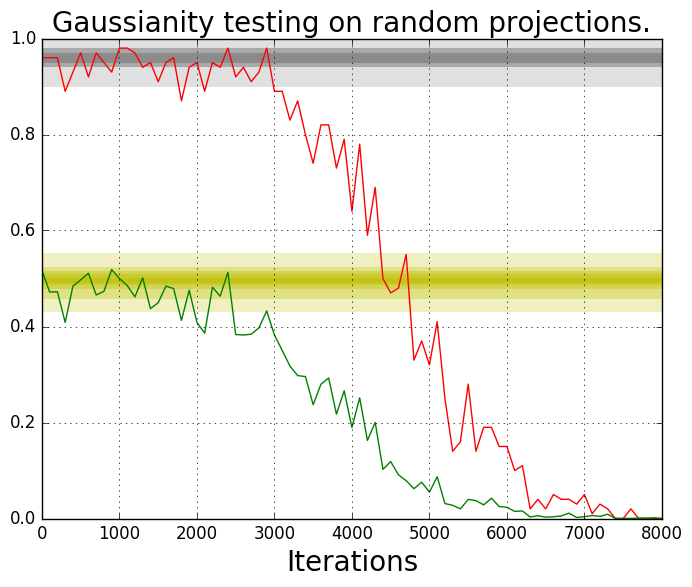}
		\caption{}
	\end{subfigure}%
	\begin{subfigure}{.32\textwidth}
		\centering
		\includegraphics[width=\linewidth]{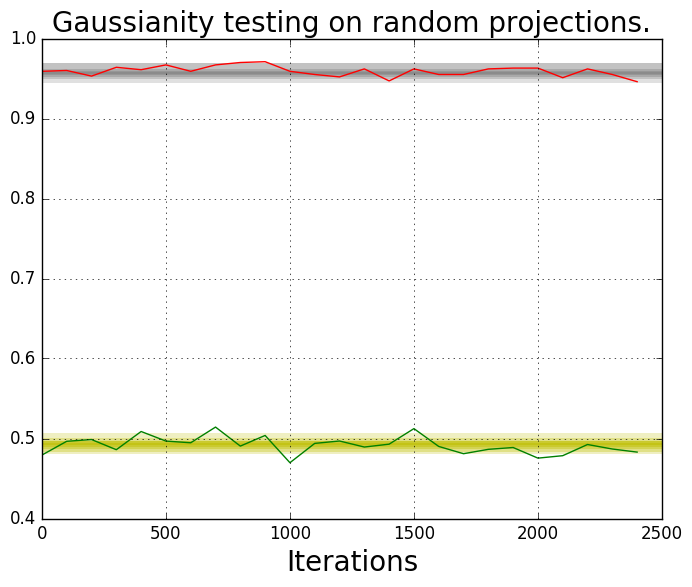}
		\caption{}
	\end{subfigure}

	\begin{flushright}
		\begin{subfigure}{.60\textwidth}
			\includegraphics[width=\linewidth]{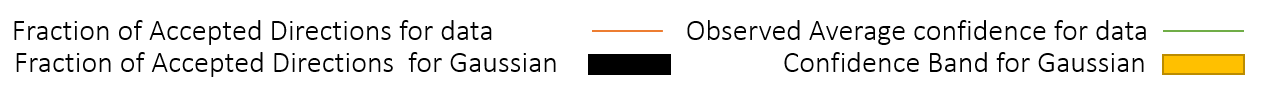}
		\end{subfigure}
	\end{flushright}
	\caption{Gaussianity Test Experiments on Resnet18, with BN and trained on CIFAR10 at l.r. $10^{-2}$, (a) Mini batch-size is $32$ (b) Mini batch-size is $256$ (c) Mini batch-size is $4096$.} 
	
\end{figure}

\begin{figure}
	\centering
	\begin{subfigure}{.32\textwidth}
		\centering
		\includegraphics[width=\linewidth]{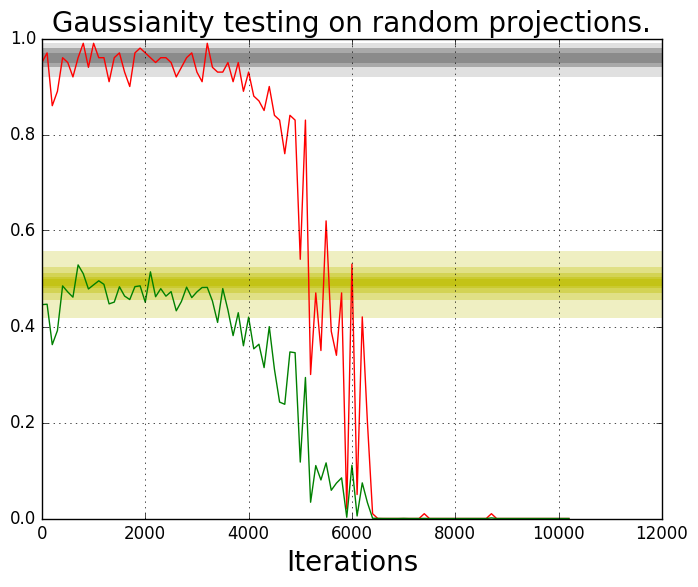}
		\caption{}
	\end{subfigure}
	\begin{subfigure}{.32\textwidth}
		\centering
		\includegraphics[width=\linewidth]{VGG16_stats/woBN_1e-2/Projection/Projection_gausstesting.png}
		\caption{}
	\end{subfigure}%
	\begin{subfigure}{.32\textwidth}
		\centering
		\includegraphics[width=\linewidth]{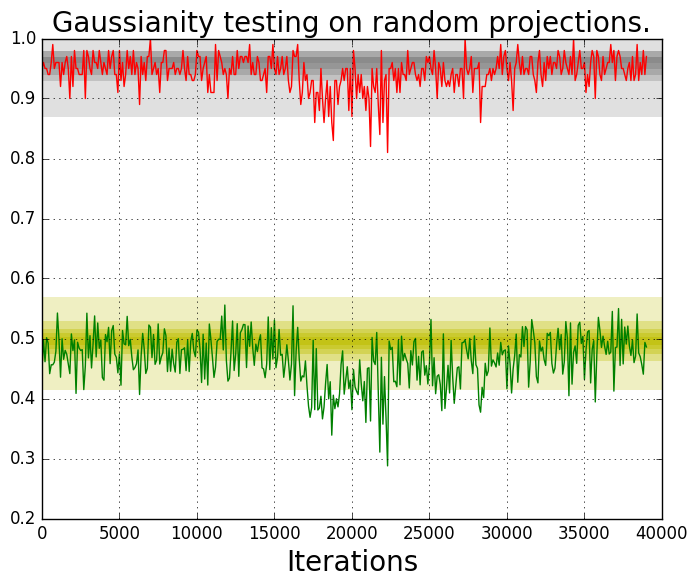}
		\caption{}
	\end{subfigure}

	\begin{flushright}
		\begin{subfigure}{.60\textwidth}
			\includegraphics[width=\linewidth]{Legend_images.png}
		\end{subfigure}
	\end{flushright}
	\caption{Gaussianity Test Experiments on VGG16, without BN and trained on CIFAR10 with $256$ minibatch-size SGD, (a) learning rate is $10^{-1}$ (b) learning rate is $10^{-2}$ (c) learning rate is $10^{-3}$.} 
	
\end{figure}

\begin{figure}
	\centering
	\begin{subfigure}{.32\textwidth}
		\centering
		\includegraphics[width=\linewidth]{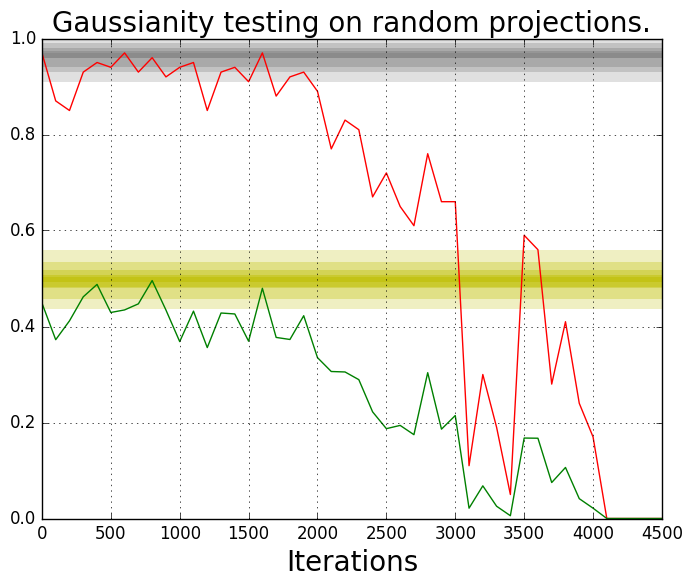}
		\caption{}
	\end{subfigure}
	\begin{subfigure}{.32\textwidth}
		\centering
		\includegraphics[width=\linewidth]{VGG16_stats/BN_1e-2/Projection/Projection_gausstesting.png}
		\caption{}
	\end{subfigure}%
	\begin{subfigure}{.32\textwidth}
		\centering
		\includegraphics[width=\linewidth]{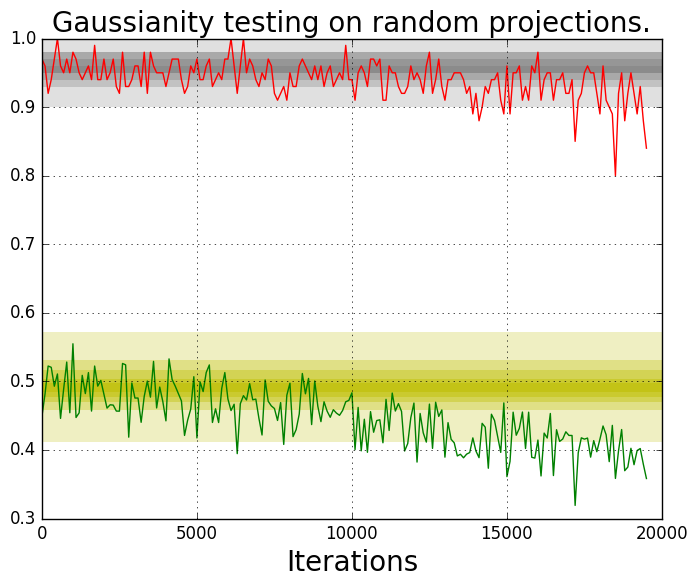}
		\caption{}
	\end{subfigure}

	\begin{flushright}
		\begin{subfigure}{.60\textwidth}
			\includegraphics[width=\linewidth]{Legend_images.png}
		\end{subfigure}
	\end{flushright}
	\caption{Gaussianity Test Experiments on VGG16, with BN and trained on CIFAR10 with $256$ minibatch-size SGD, (a) learning rate is $10^{-1}$ (b) learning rate is $10^{-2}$ (c) learning rate is $10^{-3}$.} 
	
\end{figure}

\begin{figure}
	\centering
	\begin{subfigure}{.32\textwidth}
		\centering
		\includegraphics[width=\linewidth]{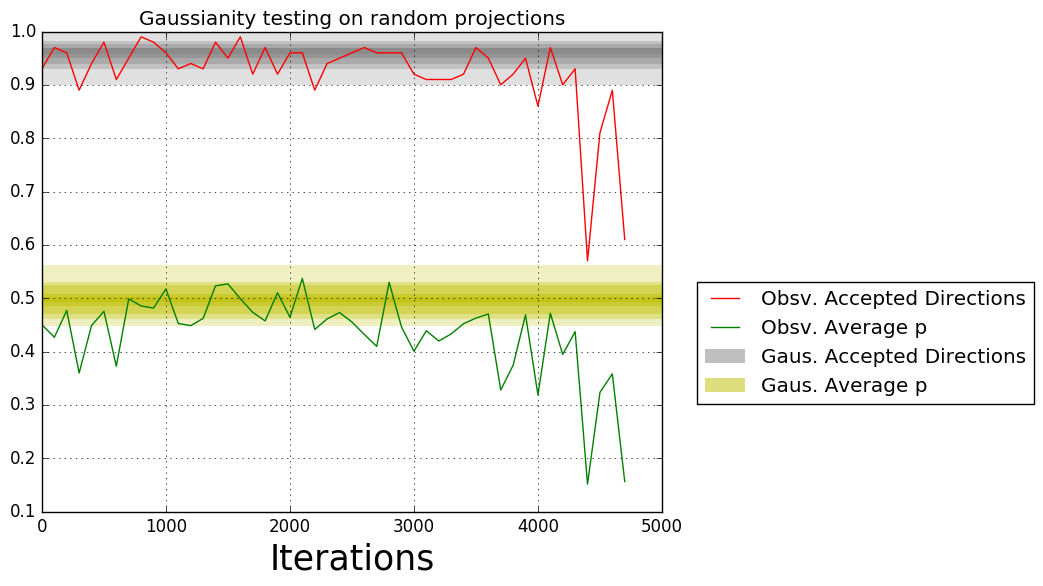}
		\caption{}
	\end{subfigure}
	\begin{subfigure}{.32\textwidth}
		\centering
		\includegraphics[width=\linewidth]{Resnet_stats/woBN1e-2/Projection/Projection_gausstesting.png}
		\caption{}
	\end{subfigure}%
	\begin{subfigure}{.32\textwidth}
		\centering
		\includegraphics[width=\linewidth]{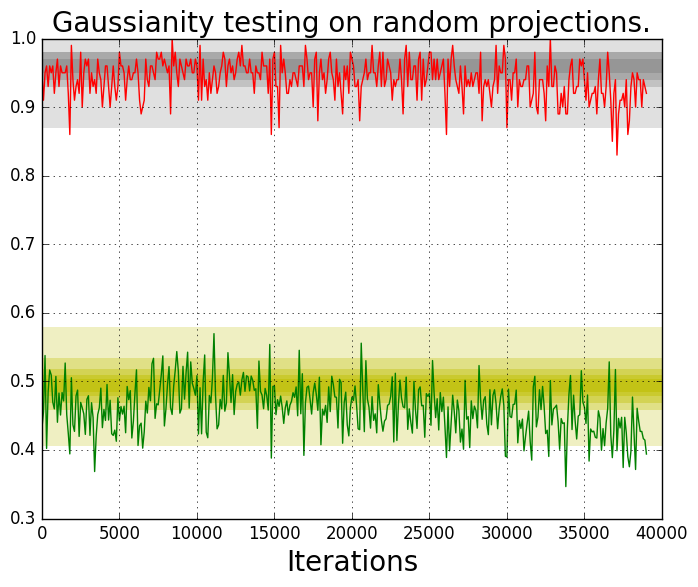}
		\caption{}
	\end{subfigure}

	\begin{flushright}
		\begin{subfigure}{.60\textwidth}
			\includegraphics[width=\linewidth]{Legend_images.png}
		\end{subfigure}
	\end{flushright}
	\caption{Gaussianity Test Experiments on Resnet18, without BN and trained on CIFAR10 with $256$ minibatch-size SGD, (a) learning rate is $10^{-1}$ (b) learning rate is $10^{-2}$ (c) learning rate is $10^{-3}$.} 
	
\end{figure}

\begin{figure}
	\centering
	\begin{subfigure}{.32\textwidth}
		\centering
		\includegraphics[width=\linewidth]{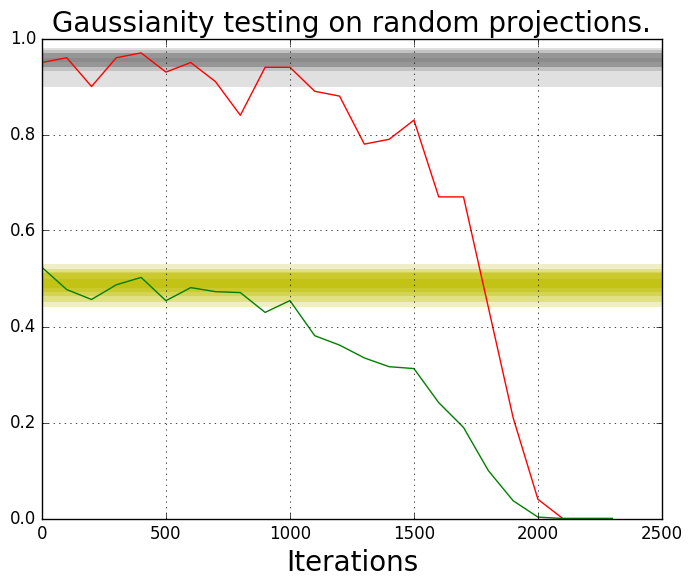}
		\caption{}
	\end{subfigure}
	\begin{subfigure}{.32\textwidth}
		\centering
		\includegraphics[width=\linewidth]{Resnet_stats/BN1e-2/Projection/Projection_gausstesting.png}
		\caption{}
	\end{subfigure}%
	\begin{subfigure}{.32\textwidth}
		\centering
		\includegraphics[width=\linewidth]{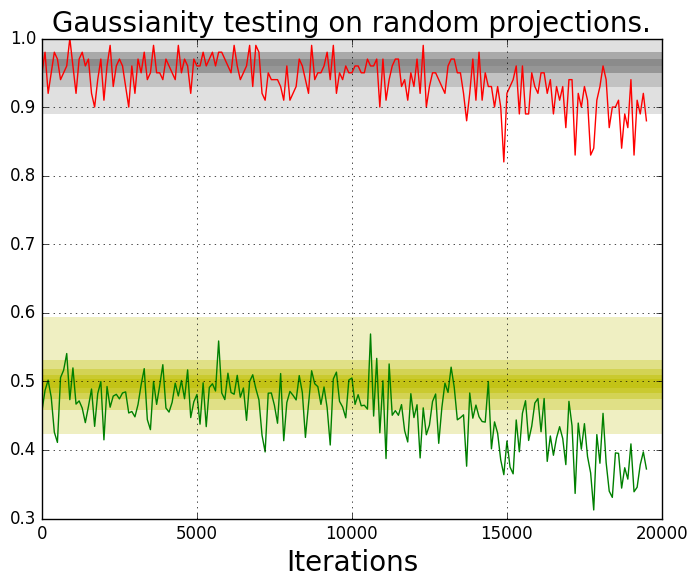}
		\caption{}
	\end{subfigure}

	\begin{flushright}
		\begin{subfigure}{.60\textwidth}
			\includegraphics[width=\linewidth]{Legend_images.png}
		\end{subfigure}
	\end{flushright}
	\caption{Gaussianity Test Experiments on Resnet18, with BN and trained on CIFAR10 with $256$ minibatch-size SGD, (a) learning rate is $10^{-1}$ (b) learning rate is $10^{-2}$ (c) learning rate is $10^{-3}$.} 
	
\end{figure}

\begin{figure}
	\centering
	\begin{subfigure}{.32\textwidth}
		\centering
		\includegraphics[width=\linewidth]{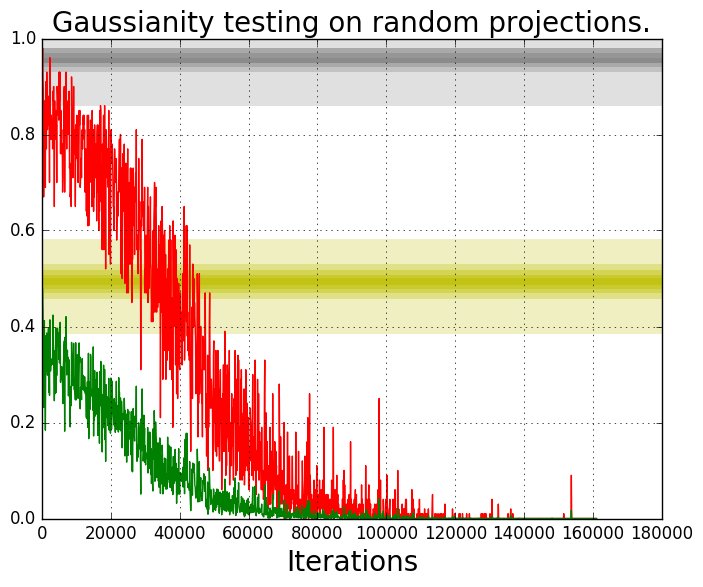}
	\end{subfigure}
	\begin{subfigure}{.32\textwidth}
		\centering
		\includegraphics[width=\linewidth]{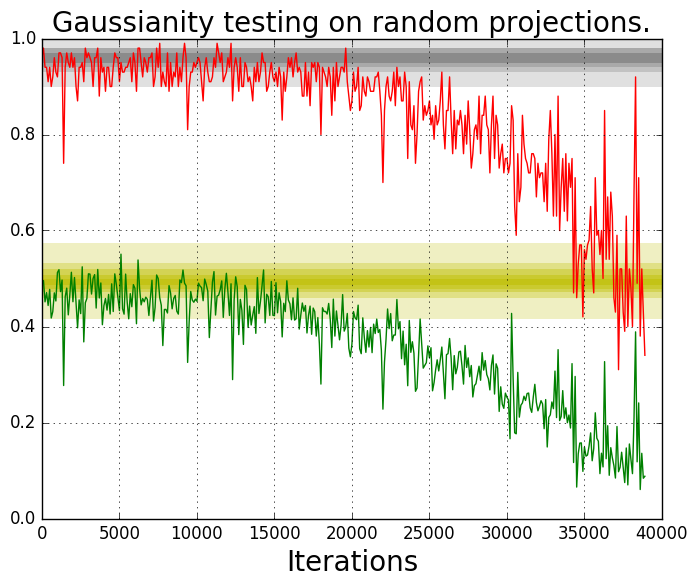}
	\end{subfigure}
	\begin{subfigure}{.32\textwidth}
		\centering
		\includegraphics[width=\linewidth]{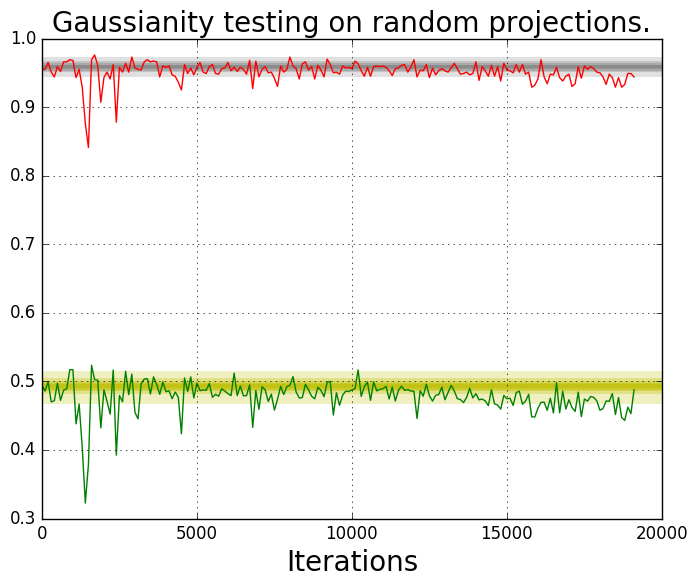}
	\end{subfigure}

	\begin{flushright}
		\begin{subfigure}{.60\textwidth}
			\includegraphics[width=\linewidth]{Legend_images.png}
		\end{subfigure}
	\end{flushright}
	\caption{Gaussianity Test Experiments on Alexnet, without BN and trained on CIFAR10 at l.r. $10^{-2}$, (a) Mini batch-size is $32$ (b) Mini batch-size is $256$ (c) Mini batch-size is $1024$.}
	
\end{figure}

\begin{figure}
	\begin{subfigure}{.32\textwidth}
		\centering
		\includegraphics[width=\linewidth]{plots_sgd_tail_index/3_0512_00_cifar10_NLL_alexnet_32_subset-1_lr0.01_gauss0_bnTrue_projectioncheck_truenbst/Projection/Projection_gausstesting.png}
		\caption{}
	\end{subfigure}
	\begin{subfigure}{.32\textwidth}
		\centering
		\includegraphics[width=\linewidth]{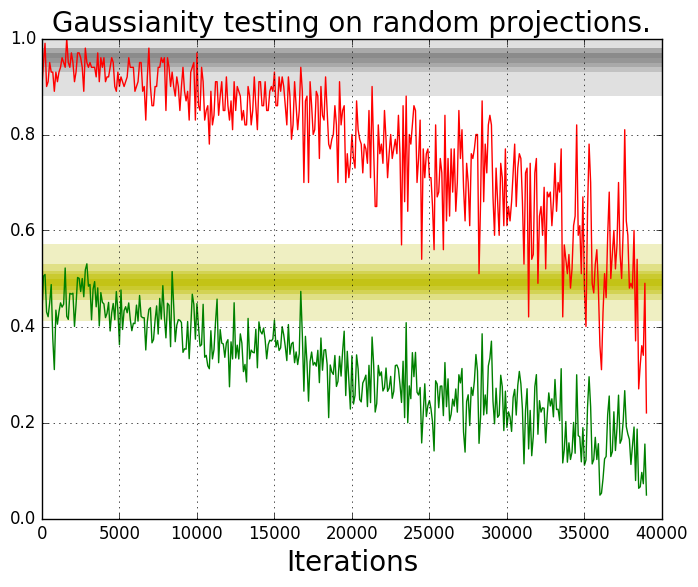}
		\caption{}
	\end{subfigure}
	\begin{subfigure}{.32\textwidth}
		\centering
		\includegraphics[width=\linewidth]{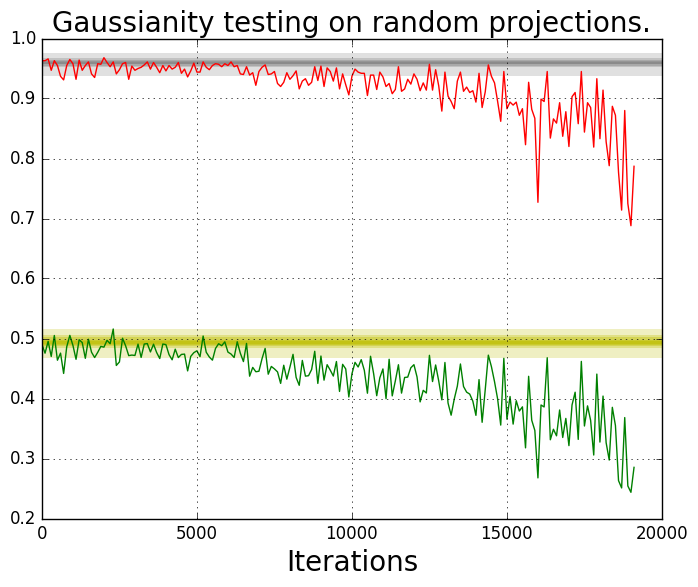}
		\caption{}
	\end{subfigure}

	\begin{flushright}
		\begin{subfigure}{.60\textwidth}
			\includegraphics[width=\linewidth]{Legend_images.png}
		\end{subfigure}
	\end{flushright}
	\caption{Gaussianity Test Experiments on Alexnet, with BN and trained on CIFAR10 at l.r. $10^{-2}$, (a) Mini batch-size is $32$ (b) Mini batch-size is $256$ (c) Mini batch-size is $1024$.}
	
\end{figure}

\begin{figure}
	\begin{subfigure}{.32\textwidth}
		\centering
		\includegraphics[width=\linewidth]{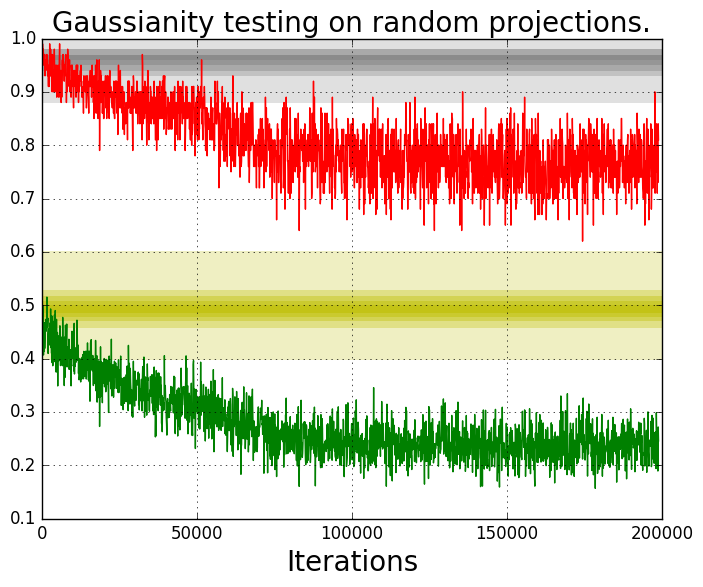}
		\caption{Gaussianity testing on gradient projection}
	\end{subfigure}
	\begin{subfigure}{.32\textwidth}
		\centering
		\includegraphics[width=\linewidth]{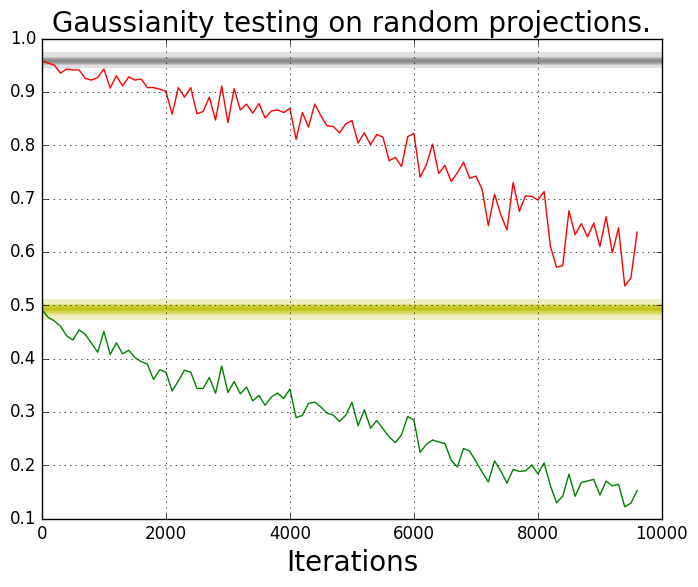}
		\caption{Gaussianity testing on gradient projection}
	\end{subfigure}
	\begin{subfigure}{.32\textwidth}
		\centering
		\includegraphics[width=\linewidth]{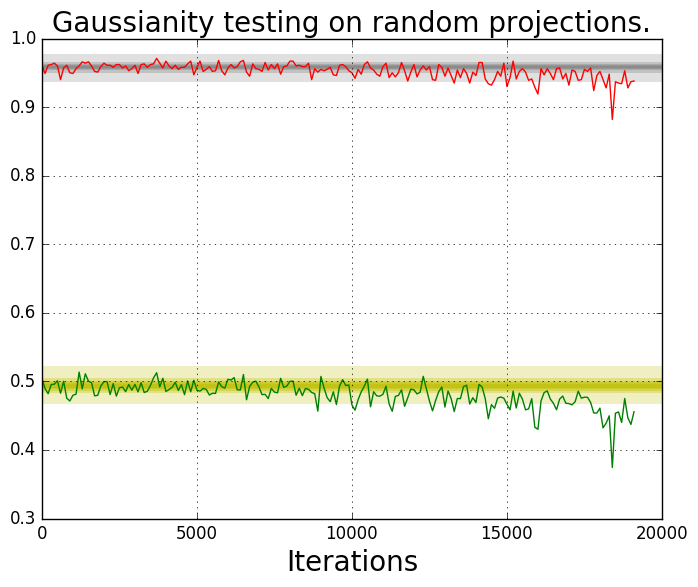}
		\caption{Gaussianity testing on gradient projection}
	\end{subfigure}

	\begin{flushright}
		\begin{subfigure}{.60\textwidth}
			\includegraphics[width=\linewidth]{Legend_images.png}
		\end{subfigure}
	\end{flushright}
	\caption{Gaussianity Test Experiments on a 3 layer fully connected network, with BN and trained on CIFAR10 at l.r. $10^{-2}$, (a) Mini batch-size is $32$ (b) Mini batch-size is $256$ (c) Mini batch-size is $1024$.}
	
\end{figure}

\begin{figure}
	\begin{subfigure}{.32\textwidth}
		\centering
		\includegraphics[width=\linewidth]{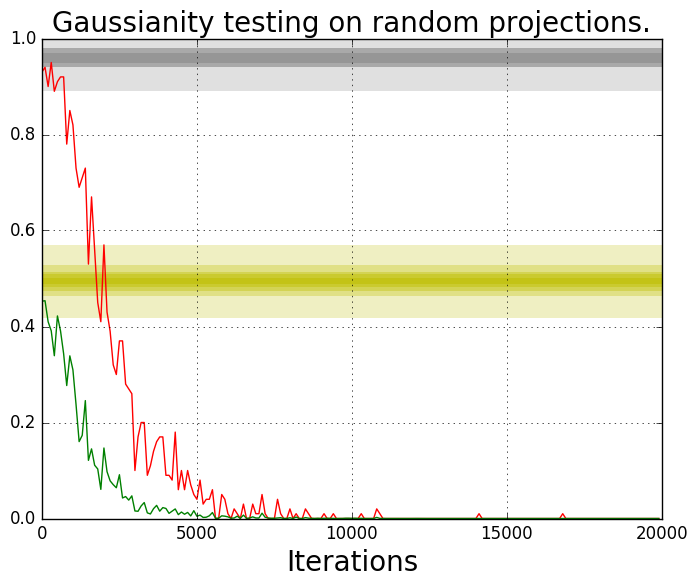}
	\end{subfigure}
	\begin{subfigure}{.32\textwidth}
		\centering
		\includegraphics[width=\linewidth]{plots_sgd_tail_index/3_0512_00_mnist_NLL_fc_256_subset-1_lr0.01_gauss0_bnFalse_projectioncheck/Projection/Projection_gausstesting.png}
	\end{subfigure}
	\begin{subfigure}{.32\textwidth}
		\centering
		\includegraphics[width=\linewidth]{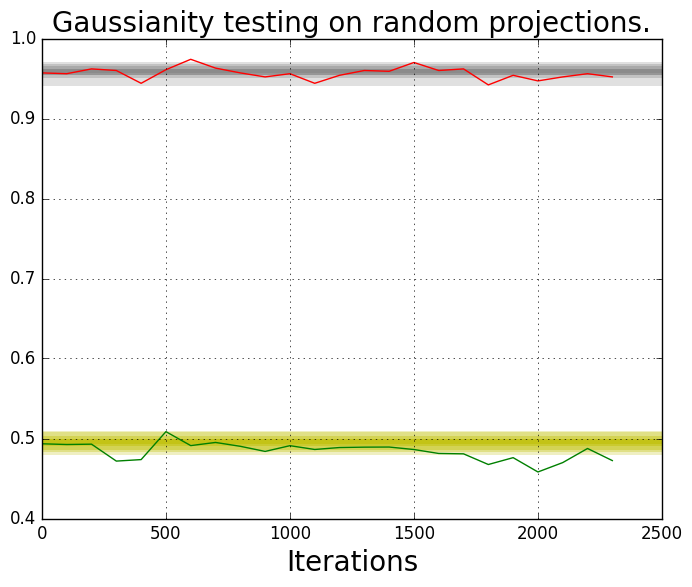}
	\end{subfigure}

	\begin{flushright}
		\begin{subfigure}{.60\textwidth}
			\includegraphics[width=\linewidth]{Legend_images.png}
		\end{subfigure}
	\end{flushright}
	\caption{Gaussianity Test Experiments on a 3 layer fully connected network, without BN and trained on MNIST at l.r. $10^{-2}$, (a) Mini batch-size is $32$ (b) Mini batch-size is $256$ (c) Mini batch-size is $1024$.}
	
\end{figure}

\begin{figure}
	\begin{subfigure}{.32\textwidth}
		\centering
		\includegraphics[width=\linewidth]{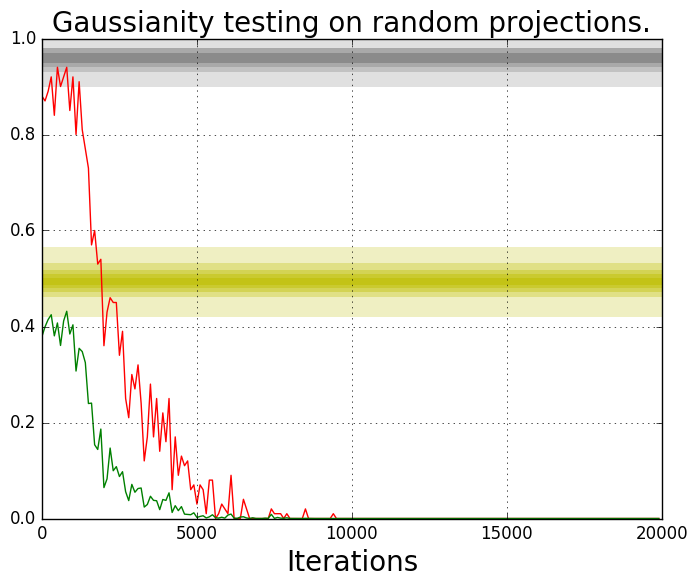}
		\caption{Gaussianity testing on gradient projection}
	\end{subfigure}
	\begin{subfigure}{.32\textwidth}
		\centering
		\includegraphics[width=\linewidth]{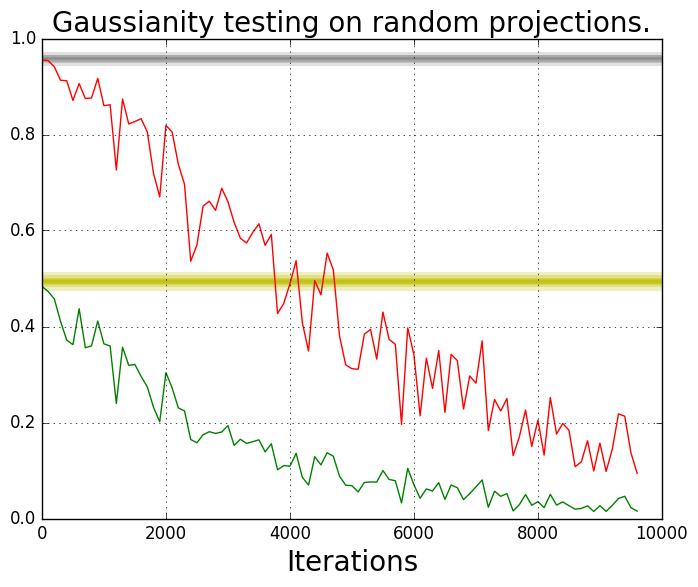}
		\caption{Gaussianity testing on gradient projection}
	\end{subfigure}
	\begin{subfigure}{.32\textwidth}
		\centering
		\includegraphics[width=\linewidth]{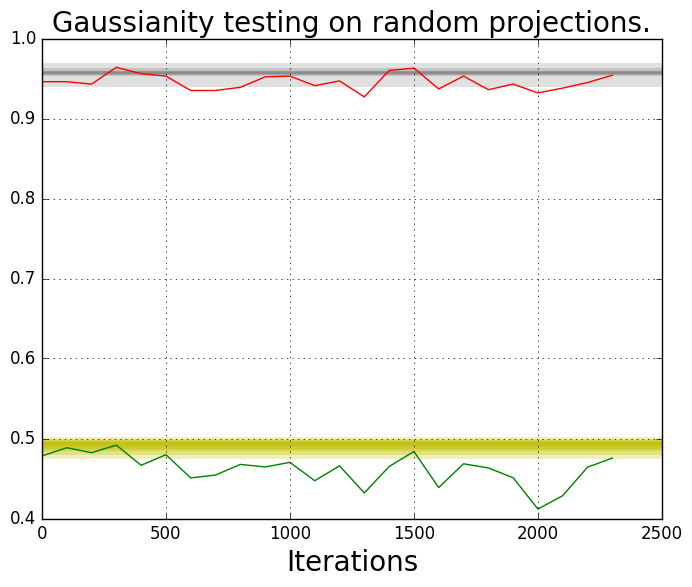}
		\caption{Gaussianity testing on gradient projection}
	\end{subfigure}

	\begin{flushright}
		\begin{subfigure}{.60\textwidth}
			\includegraphics[width=\linewidth]{Legend_images.png}
		\end{subfigure}
	\end{flushright}
	\caption{Gaussianity Test Experiments on a LeNet, without BN and trained on MNIST at l.r. $10^{-2}$, (a) Mini batch-size is $32$ (b) Mini batch-size is $256$ (c) Mini batch-size is $1024$.}
	
\end{figure}

\end{document}